\pdfoutput=1

\documentclass[11pt]{article}

\usepackage[]{EMNLP2022}

\usepackage{times}
\usepackage{latexsym}

\usepackage[T1]{fontenc}    

\usepackage[utf8]{inputenc}

\usepackage{microtype}
\usepackage{CJKutf8}
\usepackage{multirow}
\usepackage{makecell}
\usepackage{booktabs}
\usepackage{graphicx}
\usepackage{colortbl}
\usepackage{xcolor}

\usepackage{amsmath}
\usepackage{amsthm}
\usepackage{amssymb}
\usepackage{mathrsfs}
\usepackage{listings}

\definecolor{deep}{RGB}{47, 85, 151}
\definecolor{ke}{RGB}{143,170,220}

\newcommand*{\img}[1]{%
    \raisebox{-.5\baselineskip}{%
        \includegraphics[
        height=32pt,
        width=35pt,
        ]{#1}
    }
}

\title{\img{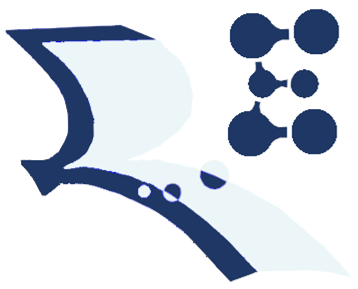}$\mathbf{\mathcal{\textcolor{deep}{D}}\textcolor{deep}{eepKE}}$: A Deep Learning Based Knowledge Extraction Toolkit for Knowledge Base Population}

\author{
Ningyu Zhang\textsuperscript{\rm 1}, 
Xin Xu\textsuperscript{\rm 1}, 
Liankuan Tao\textsuperscript{\rm 1}, 
Haiyang Yu\textsuperscript{\rm 1}, 
Hongbin Ye\textsuperscript{\rm 1}, 
Shuofei Qiao\textsuperscript{\rm 1}, \\
\textbf{Xin Xie}\textsuperscript{\rm 1}, 
\textbf{Xiang Chen}\textsuperscript{\rm 1},
\textbf{Zhoubo Li}\textsuperscript{\rm 1},
\textbf{Lei Li}\textsuperscript{\rm 1},
\textbf{Xiaozhuan Liang}\textsuperscript{\rm 1},
\textbf{Yunzhi Yao}\textsuperscript{\rm 1}, \\
\textbf{Shumin Deng}\textsuperscript{\rm 1},
\textbf{Peng Wang}\textsuperscript{\rm 1},
\textbf{Wen Zhang}\textsuperscript{\rm 1}, 
\textbf{Zhenru Zhang}\textsuperscript{\rm 2}, 
\textbf{Chuanqi Tan}\textsuperscript{\rm 2}, 
\textbf{Qiang Chen}\textsuperscript{\rm 2}, \\
\textbf{Feiyu Xiong}\textsuperscript{\rm 2}, 
\textbf{Fei Huang}\textsuperscript{\rm 2}, 
\textbf{Guozhou Zheng}\textsuperscript{\rm 1}, 
\textbf{Huajun Chen}\textsuperscript{\rm 1} \thanks{\quad Corresponding author: C.Hua (huajunsir@zju.edu.cn)}\hspace{0.5em}\\
\textsuperscript{\rm 1}  Zhejiang University \& AZFT Joint Lab for Knowledge Engine\\
\textsuperscript{\rm 2} Alibaba Group \\
 \url{http://deepke.zjukg.cn/}
}

\begin{document}
\begin{CJK*}{UTF8}{gbsn}
\maketitle
\begin{abstract}

We present an open-source and extensible knowledge extraction toolkit DeepKE, supporting complicated low-resource, document-level and multimodal scenarios in knowledge base population. DeepKE implements various information extraction tasks, including named entity recognition, relation extraction and attribute extraction. With a unified framework, DeepKE allows developers and researchers to customize datasets and models to extract information from unstructured data according to their requirements. Specifically, DeepKE not only provides various functional modules and model implementation for different tasks and scenarios but also organizes all components by consistent frameworks to maintain sufficient modularity and extensibility. We release the source code at GitHub\footnote{Github: \url{https://github.com/zjunlp/DeepKE}} with Google Colab tutorials and comprehensive documents\footnote{Docs: \url{https://zjunlp.github.io/DeepKE/}} for beginners. Besides, we present an online system\footnote{Project website: \url{http://deepke.zjukg.cn/}} for real-time extraction of various tasks, and a demo video\footnote{Video: \url{http://deepke.zjukg.cn/demo.mp4}}.

\end{abstract}

\begin{figure*}[t]
    \centering
    \includegraphics[width = 1.0\linewidth]{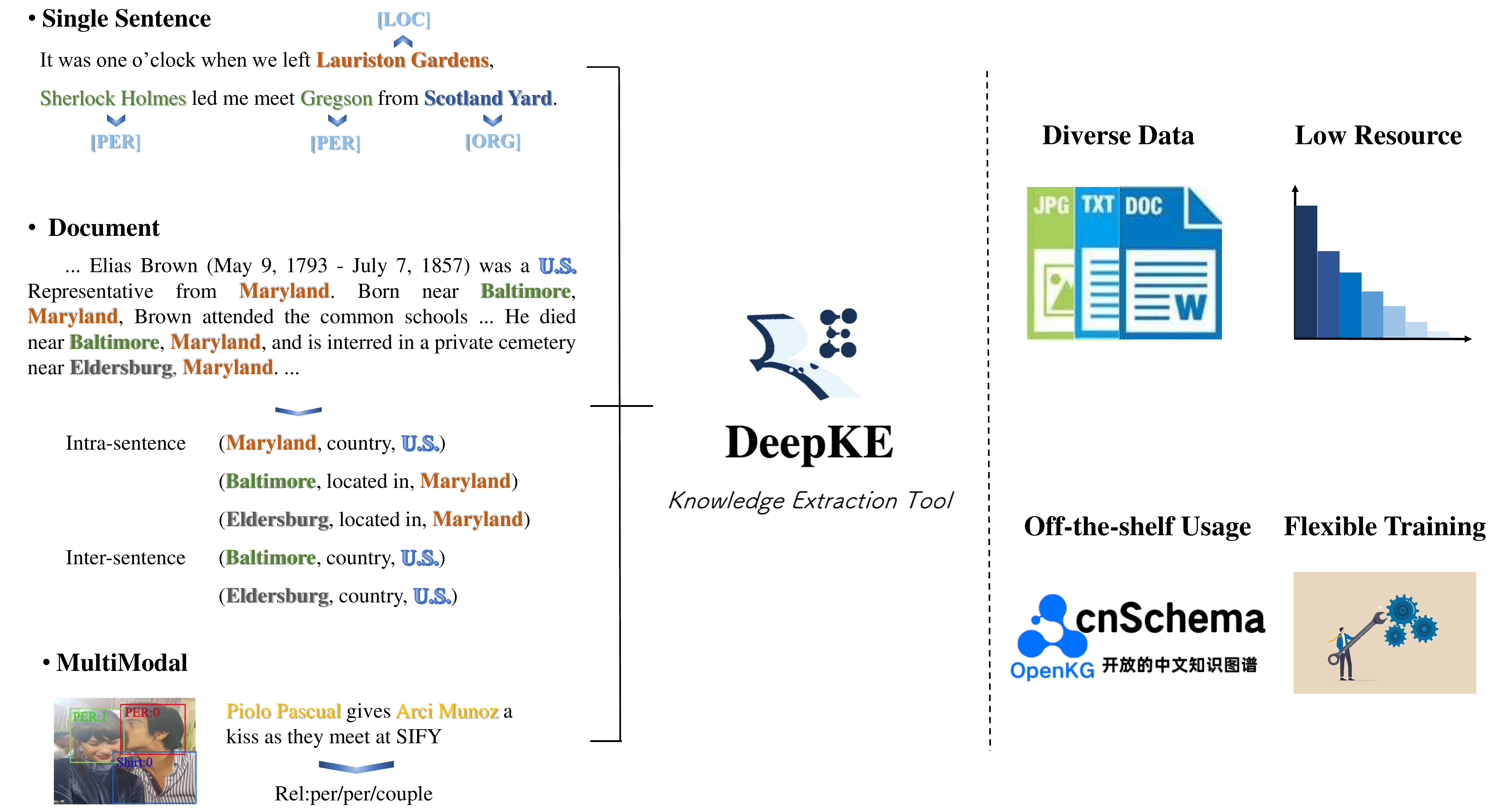}
    \captionsetup{font={normalsize}}
    \caption{The examples of tasks with different scenarios in DeepKE.}
    \label{fig:example}
\end{figure*}

\section{Introduction}


As Information Extraction (IE) techniques develop fast, many large-scale Knowledge Bases (KBs) have been constructed.
Those KBs can provide back-end support for knowledge-intensive tasks in real-world applications, such as language understanding \cite{che-etal-2021-n}, commonsense reasoning \cite{DBLP:conf/emnlp/LinCCR19} and recommendation systems \cite{DBLP:conf/www/WangZXG18}.
However, most KBs are far from complete due to the emerging entities and relations in real-world applications. 
Therefore, Knowledge Base Population (KBP) \cite{DBLP:conf/acl/JiG11} has been proposed, which aims to extract knowledge from the text corpus to complete the missing elements in KBs. 
For this target, IE is an effective technology that can extract entities and relations from raw texts and link them to KBs \cite{DBLP:conf/acl/YanGDGZQ20,DBLP:conf/emnlp/SuiW000B21}. 

To date, a few remarkable open-source and long-term maintained IE toolkits have been developed, such as Spacy \cite{vasiliev2020natural} for named entity recognition (NER), OpenNRE \cite{DBLP:conf/emnlp/HanGYYLS19} for relation extraction (RE), Stanford OpenIE \cite{DBLP:journals/eswa/Martinez-Rodriguez18} for open information extraction, RESIN for event extraction \cite{DBLP:conf/naacl/WenLLPLLZLWZYDW21} and so on \cite{jin-etal-2021-cogie}. 
However, there are still several non-trivial issues that hinder the applicability of real-world applications.

Firstly, there are various important IE tasks, but most existing toolkits only support one task.
Secondly, although IE models trained with those tools can achieve promising results, their performance may degrade dramatically when there are only a few training instances or in other complex real-world scenarios, such as encountering document-level and multimodal instances.
Therefore, it is necessary to build a knowledge extraction toolkit facilitating the knowledge base population that supports multiple tasks and complicated scenarios: \textbf{low-resource}, \textbf{document-level} and \textbf{multimodal}.

In this paper, we share with the community a new open-source knowledge extraction toolkit called \textbf{DeepKE} (MIT License), which supports knowledge extraction tasks (named entity recognition, relation extraction and attribute extraction) in the standard supervised setting and three complicated scenarios: low-resource, document-level and multimodal settings. 
To facilitate usage, we design a unified framework for data processing, model training and evaluation.
Developers and researchers can quickly customize their datasets and models for various tasks without knowing too many technical details, writing tedious glue code, or conducting hyper-parameter tuning.
We will provide maintenance to meet new requests, add new tasks, and fix bugs in the future.
We highlight our major contributions as follows:
\begin{itemize}
    \item We develop and release a knowledge base population toolkit that supports low-resource, document-level and multimodal information extraction.  
    
    \item We offer flexible usage of the toolkit with sufficient modularity as well as automatic hyper-parameter tuning; thus, developers and researchers can implement customized models for information extraction.

    \item We provide detailed documentation, Google Colab tutorials, an online real-time extraction system and long-term technical support.
\end{itemize}

\section{Core Functions}
\label{sec:app}
DeepKE is designed for different knowledge extraction tasks, including named entity recognition, relation extraction and attribute extraction. 
As shown in Figure \ref{fig:example},  DeepKE supports diverse IE tasks in standard single-sentence supervised, low-resource few-shot, document-level and multimodal settings, which makes it flexible to adapt to practical and complicated application scenarios.

\subsection{Named Entity Recognition}
As an essential task of IE, named entity recognition (NER) picks out the entity mentions and classifies them into pre-defined semantic categories given plain texts. 
For instance, given the sentence ``It was one o'clock when we left Lauriston Gardens, and Sherlock Holmes led me meet Gregson from Scotland Yard.'', NER models will predict that ``Lauriston Gardens'' as a location, ``Sherlock Holmes'' and ``Gregson'' as persons, and ``Scotland Yard'' as an organization.
To achieve supervised NER, DeepKE adopts the pre-trained language model \cite{BERT} to encode sentences and make predictions.
DeepKE also implements NER in the few-shot setting (including in-domain and cross-domain) \cite{lightner21} and the multimodal setting.

\subsection{Relation Extraction}
Relation Extraction (RE), a common task in IE for knowledge base population, predicts semantic relations between pairs of entities from unstructured texts \cite{DBLP:conf/aaai/WuLLHQZX21}. 
To allow users to customize their models, we adopt various models to accomplish standard supervised RE, including CNN \cite{DBLP:conf/emnlp/ZengLC015}, RNN \cite{zhou-etal-2016-attention}, Capsule \cite{DBLP:conf/emnlp/ZhangDSCZC18}, GCN \cite{DBLP:conf/emnlp/Zhang0M18,DBLP:conf/naacl/ZhangDSWCZC19}, Transformer \cite{DBLP:conf/nips/VaswaniSPUJGKP17} and BERT \cite{BERT}. 
Meanwhile, DeepKE provides few-shot and document-level support for RE. 
For low-resource RE, DeepKE re-implements\footnote{The code is re-organized in a unified format for flexible usage in DeepKE.} {\it KnowPrompt} \cite{KnowPrompt}, a recent well-performed few-shot RE method based on prompt-tuning.
Note that few-shot RE is significant for real-world applications, which enables users to extract relations with only a few labeled instances.
For document-level RE, DeepKE re-implements \emph{DocuNet} \cite{docunet} to extract inter-sentence relational triples within one document.
Document-level RE is a challenging task that requires integrating information within and across multiple sentences of a document \cite{DBLP:conf/acl/NanGSL20}. 
RE is also implemented in the multimodal setting described in Section \ref{sec:multimodal}.

\subsection{Attribute Extraction}
Attribute extraction (AE) plays an indispensable role in the knowledge base population. 
Given a sentence, entities and queried attribute mentions, AE will infer the corresponding attribute type. 
For instance, given a sentence ``诸葛亮，字孔明，三国时期杰出的军事家、文学家、发明家。'' (Liang Zhuge, whose courtesy name was Kongming, was an extraordinary strategist, litterateur and inventor in the Three Kingdoms period.), an entity ``诸葛亮'' (Liang Zhuge), and an attribute mention ``三国时期'' (Three Kingdoms period), DeepKE can predict the corresponding attribute type ``朝代'' (Dynasty).
DeepKE adopts various models for AE (Table \ref{tab:sdm}). 

\section{Toolkit Design and Implementation}
We introduce the design principle of DeepKE as follows:
1) \textbf{Unified Framework}: DeepKE utilizes the same framework for various task objectives with respect to \emph{Data}, \emph{Model} and \emph{Core} components;
2) \textbf{Flexible Usage}: DeepKE offers convenient training and evaluation with auto-hyperparameter tuning and the docker for operational efficiency; 
3) \textbf{Off-the-shelf Models}: DeepKE provides pre-trained models (Chinese models with pre-defined schemas) for information extraction.
We will introduce details of components in DeepKE and the unified framework in the following sections.

\begin{figure}[t!]
    \centering
    \includegraphics[width=0.48\textwidth]{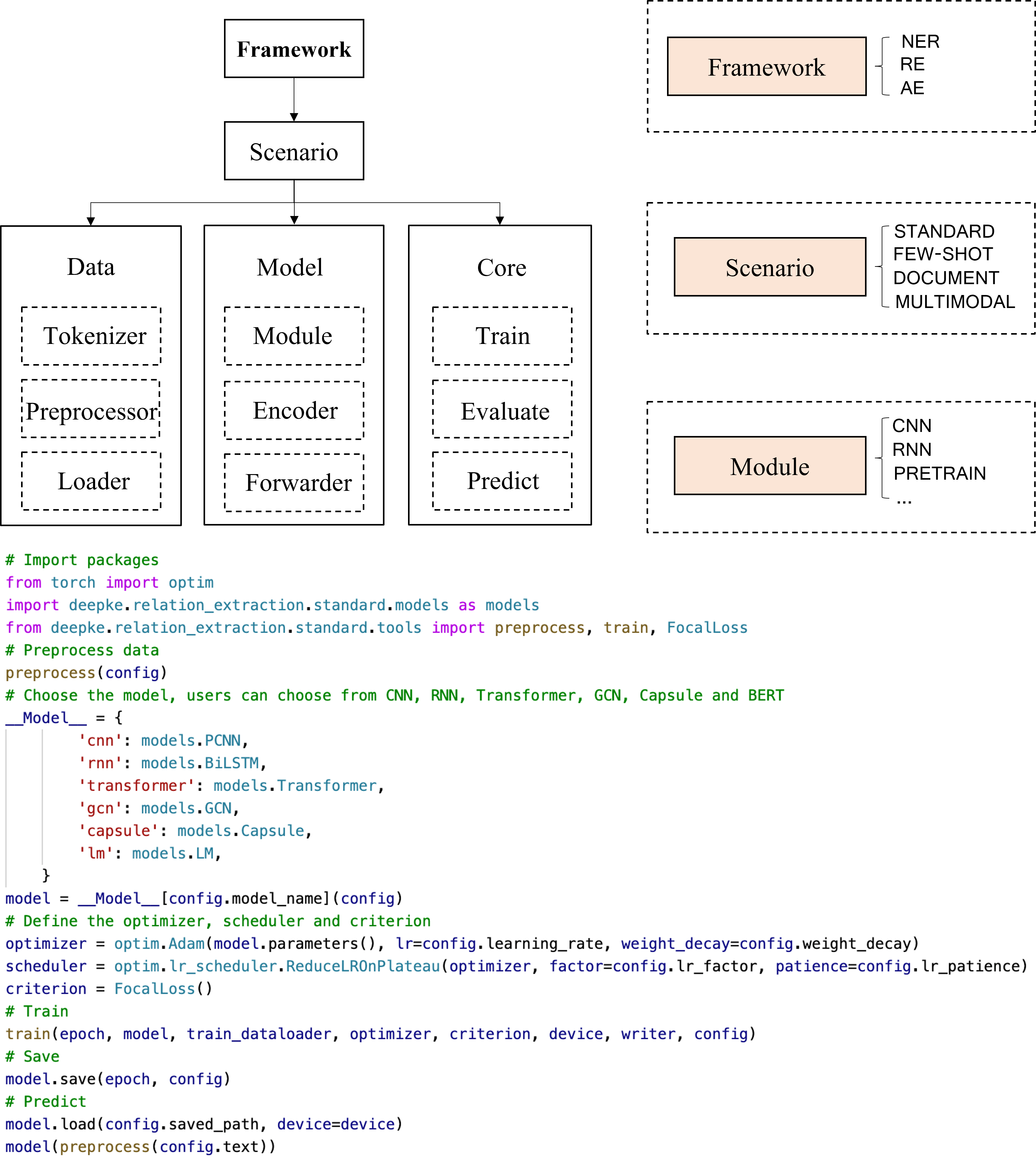}
    \caption{The architecture and example code.}
    \label{fig:deepke}
\end{figure}

\subsection{Data Module}
The data module is designed for preprocessing and loading input data. 
The tokenizer in DeepKE implements tokenization for both English and Chinese (in Appendix \ref{sec:language}).
Global images and local visual objects are preprocessed as visual information in the multimodal setting.
Developers can feed their own datasets into the tokenizer and preprocessor through the dataloader to obtain the tokens or image patches.

\subsection{Model Module}
The model module contains main neural networks leveraged to achieve three core tasks.
Various neural networks, including CNN, RNN, Transformer and the like, can be utilized for model implementation, which encodes texts into specific embedding for corresponding tasks.
To adapt to different scenarios, DeepKE utilizes diverse architectures in distinct settings, such as {\it BERT} for standard RE and {\it BART} \cite{DBLP:conf/acl/LewisLGGMLSZ20} for few-shot NER. 
We implement the \texttt{BasicModel} class with a unified \texttt{model loader} and \texttt{saver} to integrate multifarious neural models.

\subsection{Core Module}
In the core code of DeepKE, \texttt{train}, \texttt{validate}, and \texttt{predict} methods are pivotal components. 
As for the \texttt{train} method, users can feed the expected parameters (e.g., the model, data, epoch, optimizer, loss function, .etc.) into it without writing tedious glue code. 
The \texttt{validate} method is for evaluation. 
Users can modify the sentences in the configuration for prediction and then utilize the \texttt{predict} method to obtain the result.

\subsection{Framework Module}
The framework module integrates three aforementioned components and different scenarios. 
It supports various functions, including data processing, model construction and model implementation. 
Meanwhile, developers and researchers can customize all hyper-parameters by modifying configuration files formatted as {\it ``*.yaml''}, from which we apply {\it Hydra}{\footnote{\url{https://hydra.cc/}}} to obtain users' configuration.
We also offer an off-the-shelf automatic hyperparameter tuning component. 
In DeepKE, we have implemented frameworks for all application functions mentioned in Section~\ref{sec:app}. 
For other future potential application functions, we have reserved interfaces for their implementation.

\section{Toolkit Usage}

\subsection{Single-sentence Supervised Setting}
All tasks, including NER, RE and AE, can be implemented in the standard single-sentence supervised setting by DeepKE.
Every instance in datasets only contains one sentence.
The datasets of these tasks are all annotated with specific information, such as entity mentions, entity categories, entity offsets, relation types and attributes.

\subsection{Low-resource Setting}
In real-world scenarios, labeled data may not be sufficient for deep learning models to make predictions for satisfying users' specific demands.
Therefore, DeepKE provides low-resource few-shot support for NER and RE, which is exceedingly distinctive. 
DeepKE offers a generative framework with prompt-guided attention to achieve in-domain and cross-domain NER. 
Meanwhile, DeepKE implements knowledge-informed prompt-tuning with synergistic optimization for few-shot relation extraction.

\subsection{Document-Level Setting}
Relations between two entities not only emerge in one sentence but appear in different sentences within the whole document. 
Compared to other IE toolkits, DeepKE can extract inter-sentence relations from documents, which predicts an entity-level relation matrix to capture local and global information.

\subsection{Multimodal Setting}
\label{sec:multimodal}
Multimodal knowledge extraction is supported in DeepKE.
Intuitively, rich image signals related to texts are able to enhance context knowledge and help extract knowledge from complicated scenarios.
DeepKE provides a Transformer-based multimodal entity and relation extraction method named \textit{IFAformer} with prefix-based attention for multimodal NER and RE.
Specifically, \textit{IFAformer} simultaneously concatenates the textual and visual features in keys and values of the multi-head attention at each transformer layer, which can implicitly align multimodal features between texts and objects in text-related images\footnote{Implementation details in \url{https://github.com/zjunlp/DeepKE/tree/main/example/ner/multimodal}.}.

\begin{figure}[t]
    \centering
    \includegraphics[width = 1.0\linewidth]{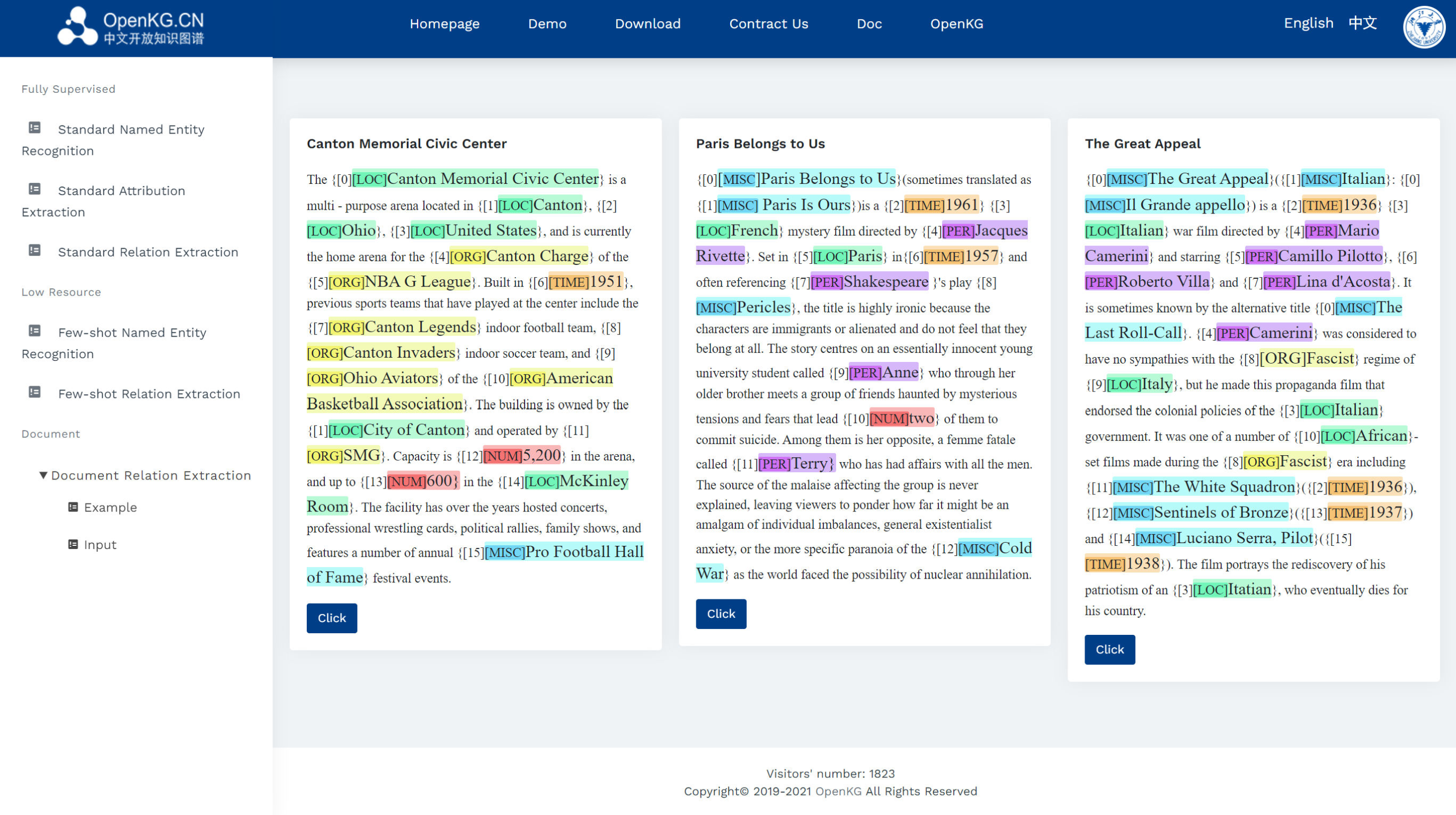}
    \captionsetup{font={normalsize}}
    \caption{An example of the online system.}
    \label{fig:online}
\end{figure}

\subsection{Online System \& \textit{cnSchema}-based Off-the-shelf Models}
Besides this toolkit, we release an online system in \url{http://deepke.zjukg.cn}.
As shown in Figure \ref{fig:online}, we train our models in different scenarios with multilingual support (English and Chinese) and deploy them for online access. 
The system can be directly applied to recognize named entities, extract relations, classify attributes from plain texts, and visualizes extracted relational triples as knowledge graphs.
The models are trained \textbf{with the pre-defined schema} (The system cannot extract knowledge out of the schema scope.) and offer flexible usage for users to obtain their customized models with their own schemas.
Furthermore, DeepKE provides off-the-shelf extraction models with Chinese pre-trained language models \cite{DBLP:journals/taslp/CuiCLQY21} based \textit{cnSchema}\footnote{\url{http://cnschema.openkg.cn/}} supporting 28 entity types and 50 relation categories.

\section{Experiment and Evaluation}


\begin{table}
  \centering
  \scalebox{0.62}{
    \begin{tabular}{ccclc}
    \toprule
    \textbf{Scenario} & \textbf{Task} & \textbf{Dataset} & \multicolumn{1}{c}{\textbf{Method}} & \textbf{F1} \\
    \midrule
    \multicolumn{1}{c}{\multirow{14}[6]{*}{\textbf{Single-sentence}}} & \multirow{2}[2]{*}{NER} & CoNLL-2003 & \multirow{2}[2]{*}{BERT} & 94.73  \\
      &   & People's Daily &   & 95.62  \\
\cmidrule{2-5}      & \multirow{6}[2]{*}{RE} & \multirow{6}[2]{*}{DuIE} & CNN & 96.74  \\
      &   &   & RNN & 94.43  \\
      &   &   & Capsule & 96.23  \\
      &   &   & GCN & 96.74  \\
      &   &   & Transformer & 96.54  \\
      &   &   & BERT & 95.79  \\
\cmidrule{2-5}      & \multirow{6}[2]{*}{AE} & \multirow{6}[2]{*}{Online} & CNN & 94.16  \\
      &   &   & RNN & 93.06  \\
      &   &   & Capsule & 94.57  \\
      &   &   & GCN & 94.50  \\
      &   &   & Transformer & 94.15  \\
      &   &   & BERT & 99.03  \\
    \midrule
    \multirow{8}[4]{*}{\textbf{Document}} & \multirow{8}[4]{*}{RE} & \multirow{8}[4]{*}{DocRED} & BERT\_base* & 53.20  \\
      &   &   & CorefBERT$_\text{base}^*$ & 56.96  \\
      &   &   & ATLOP-BERT$_\text{base}^*$ & 61.30  \\
      &   &   & \textbf{DeepKE (BERT$_\text{base}$}) & \textbf{61.86} \\
\cmidrule{4-5}      &   &   & RoBERTa\_large* & 59.62  \\
      &   &   & CorefRoBERTa$_\text{large}^*$ & 60.25  \\
      &   &   & ATLOP-RoBERTa$_\text{large}^*$ & 63.40  \\
      &   &   & \textbf{DeepKE (RoBERTa$_\text{large}$}) & \textbf{64.55} \\
    \midrule
    \multirow{8}[4]{*}{\textbf{Multimodal}} & \multirow{4}[2]{*}{NER} & \multirow{4}[2]{*}{Twitter17} & AdapCoAtt-BERT-CRF$^*$ & 84.10  \\
      &   &   & ViLBERT$_\text{base}^*$ & 85.04  \\
      &   &   & UMT* & 85.31  \\
      &   &   & \textbf{DeepKE (IFAformer)} & \textbf{87.39} \\
\cmidrule{2-5}      & \multirow{4}[2]{*}{RE} & \multirow{4}[2]{*}{MNRE} & BERT+SG* & 62.80  \\
      &   &   & BERT+SG+Att$^*$ & 63.64  \\
      &   &   & MEGA* & 66.41  \\
      &   &   & \textbf{DeepKE (IFAformer)} & \textbf{81.67} \\
    \bottomrule
    \end{tabular}}
    \caption{F1 Score (\%) of the single-sentence, document-level and multimodal scenarios. * means these baselines are from other papers.}
    \label{tab:sdm}
\end{table}

\subsection{Single-sentence Supervised Setting}
The performance of the standard single-sentence supervised setting is reported in Table \ref{tab:sdm}.


\paragraph{Named Entity Recognition}
We conduct NER experiments on two datasets: CoNLL-2003 \cite{conll03} for English and People's Daily\footnote{\url{https://github.com/OYE93/Chinese-NLP-Corpus/tree/master/NER/People's\%20Daily}} for Chinese.
The English part of CoNLL-2003 contains four types of entities: persons (PER), locations (LOC), organizations (ORG) and miscellaneous (MISC).
People's Daily dataset is a Chinese dataset containing 45,518 entities classified into three categories PER, LOC and ORG. 
It is observed that DeepKE yields comparable performance with various encoders for these datasets. 
Meanwhile, DeepKE supports any English and Chinese NER datasets with BIO tags.

\paragraph{Relation Extraction}
We conduct RE experiments on the Chinese DuIE dataset\footnote{\url{http://ai.baidu.com/broad/download}} with 10 relation categorie
Each sample contains one original sentence, one head entity, one tail entity in the sentence, their offsets, and the relation between them.
We utilize six different neural networks in DeepKE for evaluation. 
Users can select models before training by changing only one hyper-parameter\footnote{The hyper-parameter \textit{-model} to select networks is in \url{https://github.com/zjunlp/DeepKE/blob/main/example/re/standard/conf/config.yaml}.}.  
We report the performance of all models in Table \ref{tab:sdm}.

\paragraph{Attribute Extraction}
The Chinese dataset for AE is from an online resource\footnote{\url{https://github.com/leefsir/triplet\_extraction}}. 
In each sample, one entity is annotated with its attribute type, value, and offset. 
Attributes in the dataset are classified into 6 categories. 
The training set contains 13,815 samples. 
The validation set contains 3,131 samples, and the test set includes 5,921 samples.
Like RE, we leverage six neural models to extract attributes from the given sentence to evaluate DeepKE.
\subsection{Low-resource Setting}
We report the performance of the low-resource setting (NER and RE) in Table \ref{tab:in-domain}, \ref{tab:cross-domain}, and \ref{tab:few-shot-re}.

\begin{table}
  \centering
  \small
  \scalebox{0.82}{
    \begin{tabular}{lccccc}
    \toprule
    \multirow{2}[4]{*}{\textbf{Model}} & \multicolumn{5}{c}{\textbf{Entity Category}} \\
\cmidrule{2-6}      & PER & ORG & LOC* & MISC* & Overall \\
    \midrule
    LC-BERT & 76.25 & 75.32 & 61.55 & 59.35 & 68.12 \\
    LC-BART & 75.70 & 73.59 & 58.70 & 57.30 & 66.82 \\
    Template. & 84.49 & 72.61 & 71.98 & 73.37 & 75.59 \\
    \textbf{DeepKE (LightNER)} & \textbf{90.96} & \textbf{76.88} & \textbf{81.57} & \textbf{82.08} & \textbf{78.97} \\
    \bottomrule
    \end{tabular}
    }
    \caption{F1 scores of in-domain low-resource NER on CoNLL-2003. * indicates low-resource entity types (100-shot).}
  \label{tab:in-domain}
\end{table}

\begin{table}
  \centering
  \small
  \scalebox{0.87}{
    \begin{tabular}{lccc}
    \toprule
    \multirow{2}[4]{*}{\textbf{Model}} & \multicolumn{3}{c}{\textbf{Dataset}} \\
\cmidrule{2-4}      & MIT Movie & MIT Restaurant & ATIS \\
    \midrule
    Neigh.Tag. & 1.4 & 3.6 & 3.4 \\
    Example. & 29.6 & 26.1 & 16.5 \\
    MP-NSP & 36.8 & 48.2 & 74.8 \\
    LC-BERT & 45.2 & 40.9 & 78.5 \\
    LC-BART & 30.4 & 11.1 & 74.4 \\
    Template. & 54.2 & 60.3 & 88.9 \\
    \textbf{DeepKE (LightNER)} & \textbf{75.6} & \textbf{67.4} & \textbf{89.4} \\
    \bottomrule
    \end{tabular}
    }
    \caption{F1 scores of cross-domain few-shot NER (20-shot).}
  \label{tab:cross-domain}
\end{table}

\paragraph{Named Entity Recognition}
We conduct experiments in both in-domain and cross-domain few-shot settings with LightNER \cite{lightner21}.
Following \citet{cui-etal-2021-template}, for the in-domain few-shot scenario, we reduce the number of training samples for certain entity categories by downsampling one dataset.
Specifically, from CoNLL-2003, we choose \textbf{100} ``LOC'' and \textbf{100} ``MISC'' as the low-resource entities and 2,496 ``PER'' and 3,763 ``ORG'' as the rich-resource entities.
We leverage DeepKE to carry out the few-shot experiments and adopt BERT and BART with label-specific classifier layers as strong baselines denoted as {\it LC-BERT} and {\it LC-BART}.
We also use template-based BART {\it(Template.)} \cite{cui-etal-2021-template} as the competitive few-shot baseline. 
From Table \ref{tab:in-domain}, DeepKE outperforms other methods for both rich- and low-resource entity types, which illustrates that DeepKE has an outstanding performance on in-domain few-shot NER. 
In the cross-domain setting where the target entity categories and textual style are different from the source domain with limited labeled data available for training, we adopt the CoNLL-2003 dataset as an ordinary domain, and MIT Movie Review \cite{mit-dataset}, MIT Restaurant Review \cite{mit-dataset} and Airline Travel Information Systems (ATIS) \cite{atis} datasets as target domains.
The few-shot NER model in DeepKE is trained on CoNLL-2003 and fine-tuned on 20-shot target domain datasets (randomly sampled per entity category). 
We employ prototype-based {\it Neigh.Tag.} {\cite{DBLP:conf/acl/WisemanS19}}, {\it Example.} (example-based NER) \cite{example-ner},  {\it MP-NSP} (Multi-prototype+NSP) \cite{huang2020fewshot}, {\it LC-BERT}, {\it LC-BART} and {\it Template.} as competitive baselines.
From Table \ref{tab:cross-domain}, we notice that DeepKE achieves the most excellent few-shot performance.

\paragraph{Relation Extraction}
For few-shot relation extraction, we use SemEval 2010 Task-8 \cite{DBLP:conf/semeval/HendrickxKKNSPP10}, a conventional dataset of relation classification with nine bidirectional relations and one unidirectional relation \textsc{Other}.
We utilize a SOTA few-shot RE method, {\it KnowPrompt} \cite{KnowPrompt} which incorporates knowledge into prompt-tuning with synergistic optimization, to conduct 8-, 16-, and 32-shot experiments compared with other baselines, such as {\it GDPNet} \cite{DBLP:conf/aaai/XueSZC21} and {\it PTR} \cite{ptr}. Table \ref{tab:few-shot-re} shows that DeepKE outperforms those baseline methods.

\begin{table}
  \centering
  \small
    \begin{tabular}{lccc}
    \toprule
    \multirow{2}[3]{*}{\textbf{Method}} & \multicolumn{3}{c}{\textbf{Split}} \\
\cmidrule{2-4}      & K=8 & K=16 & K=32 \\ \midrule
    Fine-Tuning & 41.3 & 65.2 & 80.1 \\
    GDPNet & 42.0 & 67.5 & 81.2 \\
    PTR & 70.5 & 81.3 & 84.2 \\
    \textbf{DeepKE (KnowPrompt)} & \textbf{74.3} & \textbf{82.9} & \textbf{84.8} \\
    \bottomrule
    \end{tabular}%
     \caption{F1 scores of few-shot relation extraction}
  \label{tab:few-shot-re}
  \vspace{-6mm}
\end{table}



\subsection{Document-level Setting}
DeepKE can extract intra- and inter- sentence relations among multiple entities within one document.
We leverage a large-scale document-level RE dataset, DocRED \cite{DBLP:conf/emnlp/YeLDLLSL20}, containing 3,053/1,000/1,000 instances for training, validation and testing, respectively. 
We use cased BERT-base and RoBERTa-large \cite{DBLP:journals/corr/abs-1907-11692} as encoders. 
Compared with BERT-based and RoBERTa-based models, including {\it Coref} \cite{DBLP:conf/emnlp/YeLDLLSL20}, and {\it ATLOP} \cite{DBLP:conf/aaai/Zhou0M021}, DeepKE appears the better or comparable performance than baselines as shown in Table \ref{tab:sdm}.

\subsection{Multimudal Setting}
We report the performance of NER and RE in the multimodal scenario in Table \ref{tab:sdm}.
\paragraph{Named Entity Recognition}
Multimodal NER experiments are conducted on Twitter-2017 \cite{lu-etal-2018-visual} including texts and images from Twitter (2016-2017).
The baselines for comparison are {\it AdapCoAtt-BERT-CRF} \cite{Zhang_Fu_Liu_Huang_2018}, ViLBERT \cite{NEURIPS2019_c74d97b0} and UMT \cite{yu-etal-2020-improving-multimodal}.
We notice DeepKE can obtain a performance improvement compared with baselines.

\paragraph{Relation Extraction}
We use MNRE \cite{9428274}, a multimodal RE dataset containing sentences and images containing 23 relation categories.
Previous SOTA models including {\it BERT+SG} \cite{10.1145/3474085.3476968}, {\it BERT+SG+Att} ({\it BERT+SG} with attention calculating semantic similarity between textual and visual graphs) and {\it MEGA} \cite{10.1145/3474085.3476968}, are leveraged for comparison.
We further observe that DeepKE yields better performance than baselines.

\section{Conclusion}
In practical application, the knowledge base population struggles with low-resource, document-level and multimodal scenarios. 
To this end, we propose DeepKE, an open-source and extensible knowledge extraction toolkit. 
We conduct extensive experiments that demonstrate the models implemented by DeepKE can achieve comparable performance compared to some state-of-the-art methods. 
Besides, we provide an online system supporting real-time extraction (\textbf{with the pre-defined schemas}) without training. 
We will offer long-term maintenance to fix bugs, solve issues, add documents (tutorials) and meet new requests.

\clearpage

\section*{Broader Impact Statement} 

As noted in \citet{manning2022human}, linguistics and knowledge-based artificial intelligence were rapidly developing, and knowledge (explicit or implicit) as potential dark matter\footnote{2082: An ACL Odyssey: The Dark Matter of Language and Intelligence} for language understanding still faces obstacles to acquisition and representation.
To this end, IE technologies that aim to extract knowledge from unstructured data can serve as valuable tools to not only govern domain resources (e.g., medical, business) but also benefit deep language understanding and reasoning ability.
Note that the proposed toolkit, DeepKE, can offer flexible usage in widespread IE scenarios with pre-trained off-the-shelf models. 
We hope to deliver the benefits of the proposed DeepKE to the natural language processing community.

\section*{Acknowledgments}

We  want to express gratitude to the anonymous reviewers for their kind comments. 
This work was supported by National Natural Science Foundation of China (No.62206246, 91846204 and U19B2027), Zhejiang Provincial Natural Science Foundation of China (No. LGG22F030011), Ningbo Natural Science Foundation (2021J190), and Yongjiang Talent Introduction Programme (2021A-156-G). 
This work was supported by Information Technology Center and State Key Lab of CAD\&CG, ZheJiang University.

\bibliography{anthology,custom}
\bibliographystyle{acl_natbib}


\appendix

\section{Toolkit Usage Details}
In this section, we introduce how to use DeepKE exhaustively.
 
\subsection{Build a Model From Scratch}
\paragraph{Prepare the Runtime Environment}
Users can clone the source code from the DeepKE GitHub repository and create a runtime environment.
There are two convenient methods to create the environment.
Users can choose to either leverage {\it Anaconda} or run the docker file provided in the repository.
Besides, all dependencies can be installed by running \texttt{pip install deepke} directly.
If developers would like to modify the source code of DeepKE, the following commands should be executed: running \texttt{python setup.py install}, modifying code and then running \texttt{python setup.py develop}.
Users can also use corresponding datasets (e.g., default or customized datasets) to obtain specific information extraction models. 
All datasets need to be downloaded or uploaded in the folder named {\it data}.

\begin{table}
  \centering
  \small
    \begin{tabular}{cc}
    \toprule
    \textbf{Word} & \textbf{Named Entity Tag} \\
    \midrule
    U.N. & B-ORG \\
    official & O \\
    Ekeus & B-PER \\
    heads & O \\
    for & O \\
    Baghdad & B-LOC \\
    . & O \\
      &  \\
    Israel & B-LOC \\
    approves & O \\
    Arafat & B-PER \\
    s & O \\
    flight & O \\
    to & O \\
    West & B-LOC \\
    Bank & I-LOC \\
    . & O \\
    \bottomrule
    \end{tabular}
    \caption{Examples of the input format for NER.}
  \label{tab:nerinput}
\end{table}

\paragraph{Named Entity Recognition}
As shown in Table \ref{tab:nerinput}, the input data files with BIO tags for standard and few-shot NER contain two columns separated by a single space.
Each word has been put on a separate line, and there is an empty line after each sentence.
The two columns represent two items: the word and the named entity tag.
Before training, all datasets with the formats mentioned above should be fed into NER models through the data loader.
Developers can implement training and evaluation by running example code {\it run.py} to obtain a fine-tuned NER model, which will be used in the prediction period.
For inference, users can run {\it predict.py} with a single sentence and obtain the output recognized entity mentions and types.

\begin{figure}[t]
    \centering
    \includegraphics[width = 1.0\linewidth]{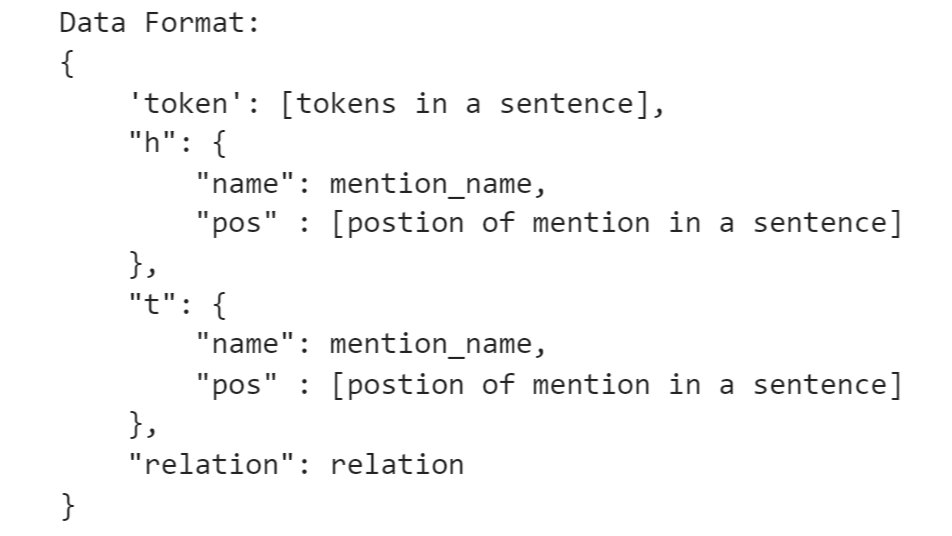}
    \captionsetup{font={normalsize}}
    \caption{The input format of few-shot RE.}
    \label{fig:few-shotREinput}
\end{figure}

\begin{figure}[t]
    \centering
    \includegraphics[width = 1.0\linewidth]{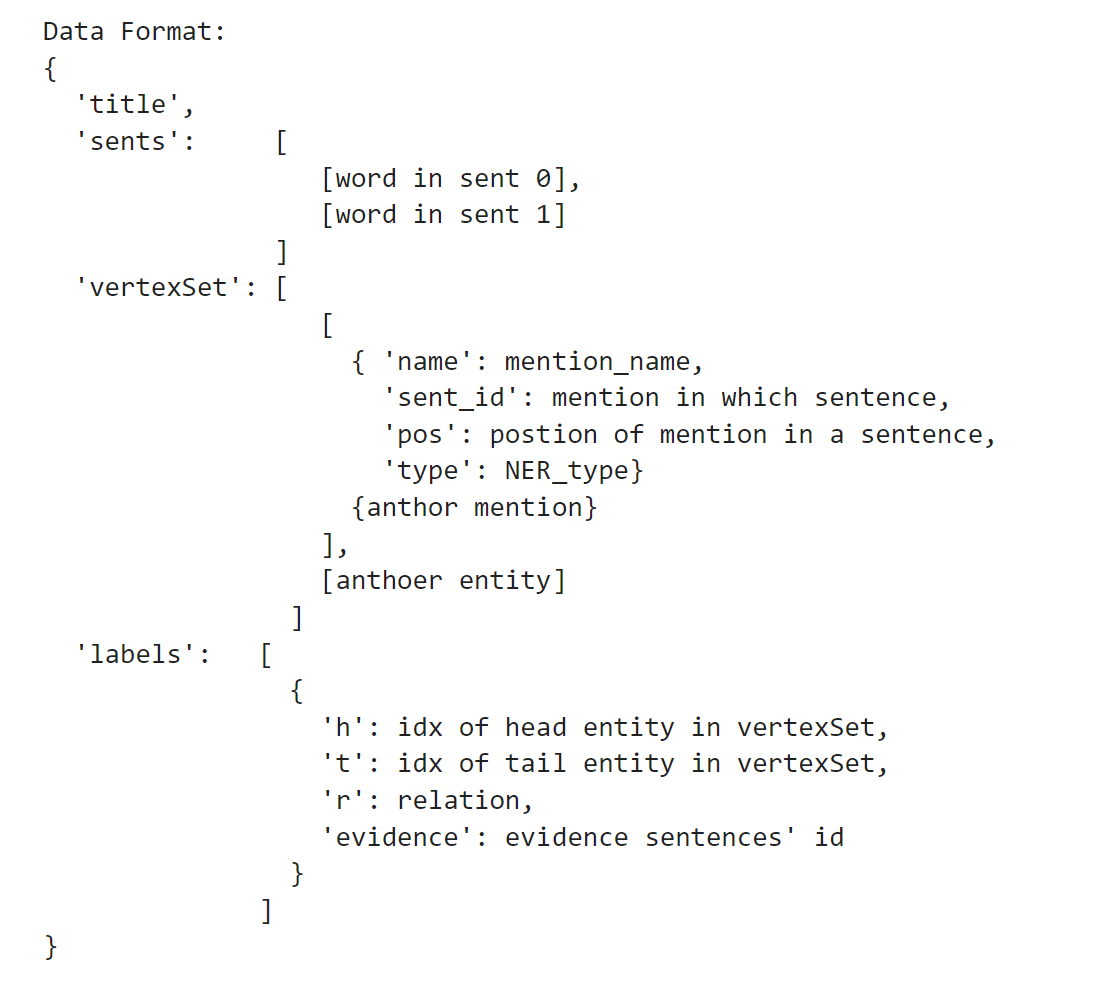}
    \captionsetup{font={normalsize}}
    \caption{The input format of document-level RE.}
    \label{fig:docREinput}
\end{figure}

\paragraph{Relation Extraction}
The training input with the CSV format of standard RE is shown in Table \ref{tab:standardRE}. 
There are five components in the format, including a sentence, a relation, the head and tail entity of the relation, the head entity offset and the tail entity offset. 
For few-shot RE, one input sample, as shown in Figure \ref{fig:few-shotREinput}, contains sentence tokens including words and punctuation, the head entity and tail entities with their mention names and position spans, and the relation between them.
For example, an input of few-shot relation extraction instance is the format of {\it \{"token": ["the", "dolphin", "uses", "its", "flukes", "for", "swimming", "and", "its", "flippers", "for", "steering", "."], "h": \{"name": "dolphin", "pos": [1, 2]\}, "t": \{"name": "flukes", "pos": [4, 5]\}, "relation": "Component-Whole(e2,e1)"\}} (h: head entity, t: tail entity, pos: position). 
The document-level RE training format is shown in Figure \ref{fig:docREinput}. 
One sample consists of a sample title, sentences separated into words and punctuation in one document, an entity set (including entity mentions, sentence IDs the entities are located in, entity position spans and entity types in the document) and a relation label set (including the head and tail entity IDs, relations and evidence sentence IDs).
After training and validation, users can run the \texttt{predict} function given an input sentence with head and tail entity to obtain corresponding relations.

\begin{table}
  \centering
  \small
   \scalebox{0.8}{
    \begin{tabular}{p{8em}p{5em}p{3em}p{1em}p{4em}p{1em}}
    \toprule
    \multicolumn{1}{c}{\textbf{Sentence}} & \multicolumn{1}{c}{\textbf{Relation}} & \textbf{Head} & \textbf{HO} & \multicolumn{1}{c}{\textbf{Tail}} & \textbf{TO} \\
    \midrule
    When it comes to beautiful sceneries in Hangzhou, West Lake first emerges in mind. & city: \newline{}located in & West Lake & 50 & \multicolumn{1}{c}{Hangzhou} & 40 \\
    \midrule
    Harry Potter, a wizard, graduated from Hogwarts School of Witchcraft and Wizardry. & school: \newline{}graduated from & Harry Potter & 0 & Hogwarts School of Witchcraft and Wizardry & 39 \\
    \bottomrule
    \end{tabular}
    }
  \caption{Examples of the input format for standard RE.\\HO: Head Offset, TO: Tail Offset.}
  \label{tab:standardRE}
\end{table}

\paragraph{Attribution Extraction}
The input CSV files formatted as Table \ref{tab:aeinput} should be given to train the attribution extraction (AE) model.
One sample contains six components: a raw sentence, a queried attribute type, an entity and its offset, the entity's corresponding attribute value and the attribute mention offset.
After training, users will obtain a fine-tuned AE model, which can be leveraged to infer attributes. 
Given a sentence with an entity and a candidate attribute mention, the AE model will predict the attribute type with confidence.
Note that all operations mentioned above are guided in the example code file {\it run.py} and {\it predict.py}.

\begin{table}
  \centering
   \scalebox{0.64}{
    \begin{tabular}{p{10em}ccccc}
    \toprule
    \multicolumn{1}{c}{\textbf{Sentence}} & \textbf{Attribute} & \textbf{Entity} & \textbf{EO} & \textbf{AV} & \textbf{AVO} \\
    \midrule
    1903年，亨利·福特\newline{}创建福特汽车公司 & 创始人 & 福特 & 9 & 亨利·福特 & 6 \\
    \midrule
    吴会期，字行可，号\newline{}子官，明朝工部郎中 & 朝代 & 吴会期 & 0 & 明朝 & 12 \\
    \bottomrule
    \end{tabular}
    }
    \caption{Examples of the input format AE. \\EO: Entity Offset, AV: Attribute Value, AVO: Attribute Value Offset.}
     \label{tab:aeinput}
\end{table}

\subsection{Auto-Hyperparameter Tuning}
To achieve automatic hyper-parameters fine-tuning, DeepKE adopts {\it Weight \& Biases}, a machine learning toolkit for developers to reduce label-intensive hyper-parameter tuning.
With DeepKE, users can visualize results and tune hyper-parameters automatically.
Note that all metrics and hyper-parameter configurations can be customized to meet diverse settings for different tasks. 
For more details on automatic hyper-parameter tuning. please refer to the official document\footnote{\url{https://docs.wandb.ai}}.


\subsection{Language Support}
\label{sec:language}
The current version of DeepKE supports English and Chinese implementation for three IE tasks, as shown in Table \ref{tab:language}.

\begin{table}
  \centering
    \begin{tabular}{cll}
    \toprule
    \textbf{Task} & \textbf{Scenario} & \textbf{Language} \\
    \midrule
    \multirow{3}[2]{*}{NER} & Supervised & Chinese \\
      & Few-shot & English, Chinese \\
      & Multimodal & English \\
    \midrule
    \multirow{4}[2]{*}{RE} & Supervised & Chinese \\
      & Few-shot & English \\
      & Multimodal & English \\
      & Document & English \\
    \midrule
    AE & Supervised & Chinese \\
    \bottomrule
    \end{tabular}
\caption{Language supported in DeepKE.}
  \label{tab:language}
\end{table}

\subsection{Notebook Tutorials}
We provide Google Colab tutorials\footnote{\url{https://colab.research.google.com/drive/1vS8YJhJltzw3hpJczPt24O0Azcs3ZpRi?usp=sharing}} and jupyter notebooks in the GitHub repository as an exemplary implementation of every task in different scenarios. These tutorials can be run directly, thus, leading developers and researchers to have a whole picture of DeepKE's powerful functions.

\section{Contributions} 
\textbf{Ningyu Zhang} from Zhejiang University, AZFT Joint Lab for Knowledge Engine, conducted the whole development of DeepKE and wrote the paper.

\textbf{Xin Xu} from Zhejiang University, AZFT Joint Lab for Knowledge Engine developed the standard NER and wrote the paper.

\textbf{Liankuan Tao, Shuofei Qiao, Peng Wang, Haiyang Yu} from Zhejiang University, AZFT Joint Lab for Knowledge Engine develop the standard RE and AE, the deepke python package, and documents and provides consistent maintenance.

\textbf{Hongbin Ye} from Zhejiang University, AZFT Joint Lab for Knowledge Engine developed the online system and constructed the online demo.

\textbf{Xin Xie, Xiang Chen} from Zhejiang University, AZFT Joint Lab for Knowledge Engine developed the few-shot relation extraction model KnowPrompt and the document-level relation extraction model DocuNet.

\textbf{Zhoubo Li, Lei Li, Xiaozhuan Liang, Yunzhi Yao, Shumin Deng, Wen Zhang} from Zhejiang University, AZFT Joint Lab for Knowledge Engine developed the Google Colab and proofread the paper.

\textbf{Zhenru Zhang, Chuanqi Tan, Qiang Chen, Feiyu Xiong, Fei Huang} from Alibaba Group, proofread the paper and advised the project.

\textbf{Guozhou Zheng, Huajun Chen} from Zhejiang University, AZFT Joint Lab for Knowledge Engine advised the project, suggested tasks, and led the research.






\end{CJK*}
\end{document}